\def\year{2019}\relax
\documentclass[letterpaper]{article} 
\usepackage{aaai19}  
\usepackage{times}  
\usepackage{helvet}  
\usepackage{courier}  
\usepackage{url}  
\usepackage{graphicx}  
\usepackage{comment}
\usepackage{subfig}
\usepackage{csquotes} 
\usepackage{devanagari} 
\usepackage{footmisc}
\usepackage{graphicx}

\makeatother

\begin{document}
%
\title{Kiki Kills: Identifying Dangerous Challenge Videos from Social Media}
\author{
Nupur Baghel\\ MIDAS Lab, NSIT-Delhi\\nupurb.co@nsit.net.in
\And Yaman Kumar \\ MIDAS Lab, IIIT-D \\yaman@nsitonline.in
\And Paavini Nanda \\ MIDAS Lab, NSIT-Delhi\\paavinin.co@nsit.net.in
\AND Rajiv Ratn Shah \\MIDAS Lab, IIIT-D\\rajivratn@iiitd.ac.in
\And Debanjan Mahata \\MIDAS Lab, IIITD \\dxmahata@ualr.edu,
\And Roger Zimmermann \\MIDAS Lab, NUS-Singapore\\rogerz@comp.nus.edu.sg
}
\maketitle
\begin{abstract}
There has been upsurge in the number of people participating in challenges made popular through social media channels. One of the examples of such a challenge is the \textit{Kiki Challenge}, in which people step out of their moving cars and dance to the tunes of the song, \emph{``Kiki, Do you love me?''}. Such an action makes the people taking the challenge prone to accidents and can also create nuisance for the others traveling on the road. In this work, we introduce the prevalence of such challenges in social media and show how the machine learning community can aid in preventing dangerous situations triggered by them by developing models that can distinguish between dangerous and non-dangerous challenge videos. Towards this objective, we release a new dataset namely \textit{MIDAS-KIKI dataset}, consisting of manually annotated dangerous and non-dangerous Kiki challenge videos. Further, we train a deep learning model to identify dangerous and non-dangerous videos, and report our results. 
\end{abstract}

\section{Introduction}
\label{introduction}
With the penetration of internet in masses, various types of uploaded content in social media have set new trends. Previous trends, like advertising \cite{miller2010social}, crisis reportage \cite{sakaki2011tweet}, dangerous selfies (\textit{killfies}) \cite{lamba:analyzing_killfie_deaths} have been thoroughly studied. There has also been some work on challenges like \emph{Blue Whale} \cite{mukhra2017blue}. However, a challenge of the magnitude and gravity such as \textit{Kiki Challenge} has not been studied previously. This is partly due to it being a very recent phenomenon, and the other reason being the unavailability of public datasets for such video-based challenges. 

The \textit{Kiki challenge}, involves people dancing to the lyrics of the song, \textit{``Kiki, Do you love me?''} and filming themselves, while a car is moving on the road. This poses a serious risk for the performers besides the people who might be using the road at the same time. Several cases of traffic blockages, injuries, accidents, crashes and even deaths have been reported relating to this challenge. Abu Dhabi, Spain and Egypt governments have started prosecution and arrest with heavy fines against the offenders{\footref{warnings}}. 

Insensitive promotion of this challenge by celebrities have been encouraging this hysteria. For instance, Will Smith performed Kiki dance on top of Budapest bridge\footnote{\url{https://www.bbc.com/news/newsbeat-44817582}}, thus effectively encouraging his 77 million followers to do the same. 
There have been arrests\footnote{\label{warnings}\url{
https://bit.ly/2NK7DLU,
https://bit.ly/2NLf4CI
%https://www.ndtv.com/mumbai-news/after-kiki-video-on-train-court-orders-mumbai-youtubers-to-clean-station-1897988 https://www.firstpost.com/entertainment/drake-releases-video-for-in-my-feelings-featuring-kiki-challenge-mastermind-shiggy-4886741.html
}}, some known and reported deaths\footnote{
\url{https://bit.ly/2MwR1T7
%https://fox40.com/2018/07/30/teen-recovering-in-icu-after-attempting-kiki-challenge/
}}, multiple cases of acute injuries\footnote{\url{
https://wapo.st/2Qx9uSR
%https://www.washingtonpost.com/news/worldviews/wp/2018/07/31/arrests-fines-and-injuries-the-in-my-feelings-challenge-has-gone-global-with-dangerous-results/
}} due to this worrying trend.

\textbf{Contributions} - In this paper, we characterize this problem, collect the relevant videos from social networking websites, release the dataset for public use and propose and build a model which can differentiate dangerous Kiki videos from the non-dangerous ones. Furthermore, we believe that our contributions would provide a tool for automatic analysis of videos such as those in Kiki challenge to reduce the number of deaths and injuries. To promote research in this domain, we also release the dataset (and metadata) that we collect for this purpose at \url{http://midas.iiitd.edu.in}. 

\begin{figure*}
\centering
\includegraphics[scale = 0.415]{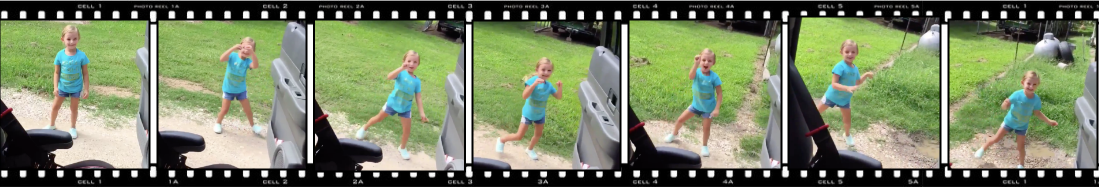}
\caption{An example for non-dangerous Kiki performance}
\label{safe_example}
\end{figure*}

\begin{figure*}
\centering
\includegraphics[scale = 0.415]{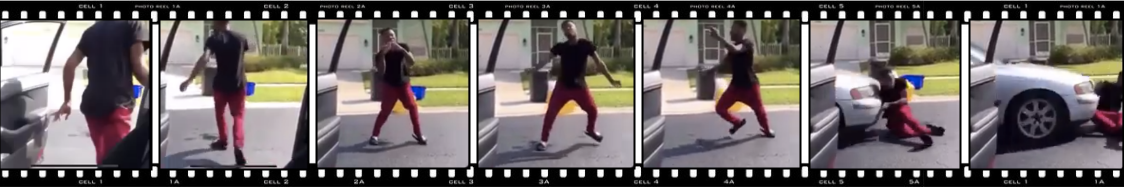}
\caption{An example for dangerous Kiki performance}
\label{unsafe_example}
\end{figure*}

\section{Methodology}
\textbf{Dataset} - We collected data from Twitter, using its publicly available APIs. We scraped the tweets and selected those containing at least one of the hashtags: \emph{\#kiki, \#keke, \#kikichallenge, \#kekechallenge, \#inmyfeelings, \#kikidoyouloveme, \#kekedoyouloveme, \#inmyfeelingschallenge, \#shiggychallenge and \#drake}. In total, we found 703 hashtags which were prevalent in tweets related to Kiki challenge. In this process, we collected more than 25k tweets, over a time period of three months from 20th June to 10th September, 2018. This period was chosen since dancing to the tunes of the song became a sensation only after the comedian, \emph{Shiggy} did the same in the later part of June.

After filtering tweets containing videos we finally obtained 2000 kiki challenge videos. Two annotators categorized these videos into \textit{dangerous} and \textit{non-dangerous}. Out of the total videos, 220 videos were tagged dangerous and the rest were non-dangerous. For categorizing a video as dangerous, the annotators were asked to answer a single question, \emph{``After watching this video, does police and/or ambulance need to be informed?"} There was a random assignment of videos to the two annotators. The inter-annotator agreement, as calculated using Cohen's Kappa was found out to be \emph{0.94}, which suggests a good agreement between the annotators. The risk perceived by annotators in most cases was from a person being hit by another car, falling on the road or getting hit by a pole while dancing, and in some cases, of overt acts of aggression by the performers, such as burning the car.  A sample of dangerous and non-dangerous video are shown in Figures \ref{safe_example} and \ref{unsafe_example}.

\begin{figure}
\centering
\includegraphics[scale = 0.40]{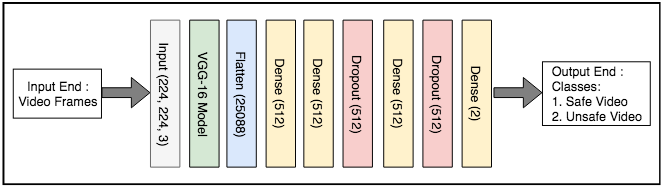}
\caption{Model architecture for classification of Kiki challenge videos.}
\label{kiki_model}
\end{figure}

\textbf{Model} - We trained a deep learning model using transfer learning on the annotated dataset. The architecture of the model is shown in Figure \ref{kiki_model}. The input video frames are fed to a VGG-16 \cite{simonyan2014very} model wrapped in a TimeDistributed wrapper in Keras in conjugation with a flattened layer followed by several dense and dropout layers (with dropout as 0.5). The last layer decides whether the video fed to the system was dangerous or not. In order to train the model, the dataset was splitted into three parts: 60\% for training, 20\% for validation and 20\% for testing using stratified sampling. 
The validation set was used for deciding the architecture and parameters of the final model. After training we tested our model on the test data, obtaining and accuracy of $87\%$. The performance metrics of the model is reported in Table \ref{performance_metric}.

\begin{table}[h!]
\centering
\caption{Performance metrics for our model.}
\label{performance_metric}
\begin{tabular}{llll} \hline
\textbf{Accuracy}&\textbf{Precision} & \textbf{Recall} & \textbf{F1-Score} \\ \hline
0.87& 0.96 & 0.90 & 0.93\\ 
\hline
\end{tabular}
\end{table}

\section{Conclusion}

In this paper, we characterized the recent trend of taking widely popular challenges on social media platforms that can create safety hazards and life-threatening situations. We took \emph{Kiki Challenge} as our use case, and implemented a deep learning model on videos that differentiates between dangerous and non-dangerous Kiki Challenge videos collected from Twitter. Our model shows high accuracy in identifying the two classes. In this process, we compiled a manually annotated dataset of dangerous and non-dangerous videos, and make it publicly available. We believe that our work would encourage researchers to use machine learning for identifying potentially dangerous challenge videos, and promote public safety. As part of our future work, we would like to investigate the problem domain in greater detail by taking into account the data imbalance and considering other metadata associated with the videos, while training our models.

\fontsize{9.0pt}{10.0pt}
\selectfont
\bibliographystyle{aaai}
\bibliography{bibliography}

\begin{thebibliography}{}

\bibitem[\protect\citeauthoryear{Lamba \bgroup et al\mbox.\egroup
  }{2016}]{lamba:analyzing_killfie_deaths}
Lamba, H.; Bharadhwaj, V.; Vachher, M.; Agarwal, D.; Arora, M.; and Kumaraguru,
  P.
\newblock 2016.
\newblock Me, myself and my killfie: Characterizing and preventing selfie
  deaths.
\newblock Technical report.

\bibitem[\protect\citeauthoryear{Miller and Lammas}{2010}]{miller2010social}
Miller, R., and Lammas, N.
\newblock 2010.
\newblock Social media and its implications for viral marketing.
\newblock {\em Asia Pacific Public Relations Journal} 11(1):1--9.

\bibitem[\protect\citeauthoryear{Mukhra \bgroup et al\mbox.\egroup
  }{2017}]{mukhra2017blue}
Mukhra, R.; Baryah, N.; Krishan, K.; and Kanchan, T.
\newblock 2017.
\newblock ‘blue whale challenge’: A game or crime?
\newblock {\em Science and engineering ethics}  1--7.

\bibitem[\protect\citeauthoryear{Sakaki, Toriumi, and
  Matsuo}{2011}]{sakaki2011tweet}
Sakaki, T.; Toriumi, F.; and Matsuo, Y.
\newblock 2011.
\newblock Tweet trend analysis in an emergency situation.
\newblock In {\em Proceedings of the Special Workshop on Internet and
  Disasters}, ~3.
\newblock ACM.

\bibitem[\protect\citeauthoryear{Simonyan and
  Zisserman}{2014}]{simonyan2014very}
Simonyan, K., and Zisserman, A.
\newblock 2014.
\newblock Very deep convolutional networks for large-scale image recognition.
\newblock {\em arXiv preprint arXiv:1409.1556}.

\end{thebibliography}
\end{document}